\documentclass[sigconf]{acmart}
\usepackage{paralist}
\usepackage{multirow}

\usepackage{color}

\setcopyright{acmcopyright}
\copyrightyear{2018}
\acmYear{2018}
\acmDOI{10.1145/1122445.1122456}

\begin{document}

\title{Exploiting Class Labels to Boost Performance on Embedding-based Text Classification}


\author{Arkaitz Zubiaga}
\orcid{0000-0003-4583-3623}
\email{a.zubiaga@qmul.ac.uk}
\affiliation{%
  \institution{Queen Mary University of London}
  \city{London, UK}
}


\begin{abstract}
  Text classification is one of the most frequent tasks for processing textual data, facilitating among others research from large-scale datasets. Embeddings of different kinds have recently become the de facto standard as features used for text classification. These embeddings have the capacity to capture meanings of words inferred from occurrences in large external collections. While they are built out of external collections, they are unaware of the distributional characteristics of words in the classification dataset at hand, including most importantly the distribution of words across classes in training data. To make the most of these embeddings as features and to boost the performance of classifiers using them, we introduce a weighting scheme, Term Frequency-Category Ratio (TF-CR), which can weight high-frequency, category-exclusive words higher when computing word embeddings. Our experiments on eight datasets show the effectiveness of TF-CR, leading to improved performance scores over the well-known weighting schemes TF-IDF and KLD as well as over the absence of a weighting scheme in most cases.
\end{abstract}




\maketitle

\section{Introduction}

Word embeddings, or distributed word representations, have become one of the most common features for processing text. Word embeddings have been successfully used in numerous NLP and IR tasks such as sentiment analysis \cite{bollegala2016cross}, machine translation \cite{zou2013bilingual}, search \cite{ganguly2015word} or recommender systems \cite{musto2016learning}, as well as different domains such as biomedicine \cite{chiu2016train} or finance \cite{cortis2017semeval}, outperforming traditional vector representation methods based on bags-of-words or n-grams.

In this work we focus on text classification \cite{sebastiani2002machine}, where word embeddings and derivatives are commonly used to represent the textual content of the instances to be classified \cite{wang2016semantic}. While word embeddings are widely used features for text classification, they are generally used for vector representation of the textual content of the instances, independent of the importance each word can have within each category. We propose to incorporate information derived from category labels in the training data to improve vector representations. Here we propose Term Frequency-Category Ratio (TF-CR), a weighting scheme that exploits the category labels from training data to produce an improved vector representation using word embeddings, which is informed by category distributions in the training data.

There is a dearth of work in the literature trying to exploit the distribution of content across classes in the training data for improving word embedding representations. The few works that tackled the problem have mainly focused on doing so for large-scale datasets, where it is possible to train separate word embedding models for each category thanks to the availability of abundant in-domain data. This is the case for problems such as sentiment analysis, where one can build such large annotated datasets by using distant supervision. In this paper, we aim to develop an improved word embedding representation for text classification where training data is not necessarily so abundant. To do that, we are the first to propose a novel weighting scheme, Term Frequency-Category Ratio (TF-CR), which can be applied on pre-trained, domain-agnostic word embedding models, only leveraging the training data available in the dataset at hand for dataset-specific weighting of the embeddings. The intuition behind TF-CR is to assign a higher weight to high-frequency, category-exclusive words as observed in the training data.

Our experiments on eight classification datasets show the consistent effectiveness of TF-CR, significantly improving performance of word embedding representations for text classification over the use of the well-known weighting schemes TF-IDF and KLD, as well as over unweighted word embeddings.

\section{Related Work}

Early methods for learning distributed representations of words \cite{bengio2003neural} through the so-called neural probabilistic language models have more recently gained momentum as embeddings \cite{pilehvar2020embeddings,grohe2020word2vec}. It sparked development of additional methods to reduce the dimensionality of traditional vector representation methods such as bags-of-words, by learning word embeddings. Two of the best-known methods to learn word embeddings include Word2Vec \cite{mikolov2013distributed} and Glove \cite{pennington2014glove}, which enable dimensionality reduction as well as capturing semantic similarities across words. The key intuition is that, having a large corpus to train a model from, one can learn semantic characteristics of words by analysing their context, i.e. words surrounding other words. This leads to vectors of reduced dimensionality (word embeddings) to represent each word, which are normally between 100 and 500 dimensions. One of the widely adopted practices for sentence representation is then to get the sum or the average of the word embeddings in the sentence in question \cite{zhang2018aggregating}.

Use of word embeddings for text classification without specific weighting of words, however, ignores potentially useful information that can be extracted from class labels. While this problem has been tackled before, there is limited work exploring the utility of class labels to make the most of word embeddings for text classification. Previous work leveraging class labels to boost the performance of word embeddings on text classification has largely focused on sentiment analysis. The sentiment analysis task is suitable as it is possible to collect large, distantly supervised datasets \cite{go2009twitter} which are exploited to train sentiment-specific embeddings. Having large annotated datasets, one can then train separate word embedding models for each class or learn models that incorporate class distributions in them. This has been achieved in different methods by combining multiple neural networks \cite{tang2014learning,tang2015sentiment,tang2016sentiment,kuang2018class,kuang2019learning} or by using separate training processes \cite{zubiaga2018learning} to train different word embedding models for each class in the dataset. This however requires availability of very large collections of labelled data to train separate models, which is possible for classification tasks exploiting distant supervision for data collection, as is the case with sentiment analysis. However, it is more limited for other text classification problems where gathering labelled data is expensive. In what follows, we propose a new weighting scheme to tackle this problem, TF-CR.

\section{The TF-CR Weighting Scheme}
\label{sec:tfcr}

We propose a novel weighting scheme for word embedding representation in text classification tasks, which aims to determine the importance of each word for each particular category based on the distribution of the word across categories in the training data; this can provide additional information that word embeddings inherently disregard. This can be achieved by using well-known weighting schemes which are often used for text classification based on bags-of-words, such as TF-IDF \cite{jones1972statistical,salton1988term} and Kullback-Leibler Divergence (KLD) \cite{kullback1951information}.

To suit the purposes of word embeddings in text classification, here we propose a new weighting scheme. The Term Frequency-Category Ratio (TF-CR) is a simple weighting scheme that combines the importance of a word within a category (Term Frequency, TF) and the distribution of the word across all categories (Category Ratio, CR). Both TF and CR are computed for each word $w$ within each category $c$. TF is computed as the ratio of words in a category that are $w$, i.e. $TF_{wc} = \frac{|w_c|}{N_c}$, where $|w_c|$ is the number of occurrences of $w$ in $c$, and $N_c$ is the total number of words in $c$. CR is computed as the ratio of occurrences of $w$ that occur within the category $c$, i.e. $CR_{wc} = \frac{|w_c|}{|w|}$, where $|w|$ denotes the number of occurrences of $w$ across all categories.

The final TF-CR is the product of both metrics (Equation \ref{eq:tfcr}).

\begin{equation}
 TF-CR = \frac{|w_c|}{N_c} * \frac{|w_c|}{|w|} = \frac{|w_c|^2}{N_c * |w|}
 \label{eq:tfcr}
\end{equation}

TF-CR ultimately gives a high weight to words that occur exclusively and with high frequency within a category. Low-frequency words exclusive to a category and high-frequency words that frequently occur across all categories will get lower scores.

\subsection{Applying TF-CR on embeddings}
\label{ssec:applying}

In order to create a representation weighted using TF-CR, we first build category-specific word embedding representations of a text. This category-specific representation is created by summing up the embeddings of each of the words in a sentence, multiplied by their TF-CR score. This leads to $k$ TF-CR-weighted embedding representations, where $k$ is the number of categories in the dataset. We finally concatenate these $k$ embedding representations to produce the final vector, which has a dimensionality of $k \times d$, where $d$ is the number of dimensions of the word embedding model.

\section{Experiments}

\subsection{Datasets}

We use eight different datasets:

\begin{compactitem}
 \item \textbf{RepLab polarity dataset \cite{replab2013overview}:} A dataset of 84,745 tweets mentioning companies, annotated for polarity as positive, negative or neutral.\footnote{\url{http://nlp.uned.es/replab2013/}}
 \item \textbf{ODPtweets \cite{zubiaga2013harnessing}:} a large-scale dataset with nearly 25 million tweets, each categorised into one of the 17 categories of the Open Directory Project (ODP).
 \item \textbf{Restaurant reviews \cite{jiang2019leveraging}:} a large dataset of 14,542,460 TripAdvisor restaurant reviews with their associated star rating ranging from 1 to 5.
 \item \textbf{SemEval sentiment tweets \cite{rosenthal2017semeval}:} we aggregate all annotated tweets from the SemEval Twitter sentiment analysis task from 2013 to 2017. The resulting dataset contains 61,767 tweets.
 \item \textbf{Distantly supervised sentiment tweets:} by using a large collection of tweets from January 2013 to September 2019 released on the Internet Archive\footnote{\url{https://archive.org/details/twitterstream}}, we produce a dataset of tweets annotated for sentiment analysis by using distant supervision following \cite{go2009twitter}, leading to tweets annotated as positive or negative. The resulting dataset contains 33,203,834 tweets.\footnote{\url{http://www.zubiaga.org/datasets/sentiment1319/}}
 \item \textbf{Hate speech dataset \cite{founta2018large}:} a dataset of 99,996 tweets, each categorised into one of \{abusive, hateful, spam, normal\}.
 \item \textbf{Newsspace200 \cite{del2005ranking}:} a dataset of nearly 500K news articles, each categorised into one of 14 categories, including business, sports entertainment.\url{http://groups.di.unipi.it/~gulli/AG_corpus_of_news_articles.html}
 \item \textbf{20 Newsgroups:} a collection of nearly 20,000 newsgroup documents, pertaining to 20 different newsgroups, which are used as categories.\footnote{\url{http://qwone.com/~jason/20Newsgroups/}}
\end{compactitem}

For all datasets, we randomly sample 100,000 instances, except for those with fewer instances.

\subsection{Word Embedding Models \& Classifiers}

We tested four word embedding models: (1) Google's Word2Vec model (\textbf{gw2v}), (2) a Twitter Word2Vec model\footnote{\url{https://fredericgodin.com/software/}} (\textbf{tw2v}) \cite{godin2015multimedia}, (3) GloVe embeddings trained from Common Crawl (\textbf{cglove}) and (4) GloVe embeddings trained from Wikipedia (\textbf{wglove}).\footnote{\url{https://nlp.stanford.edu/projects/glove/}}

Different classifiers were tested. Due to limited space, we show here results obtained with a logistic regression classifier and \textit{tw2v} embeddings, which consistently lead to optimal results. We report macro-F1 values as performance scores.

\subsection{Weighting Schemes}

We compare four different weighting schemes, all of which are applied following the methodology in \S \ref{ssec:applying}:

\begin{compactitem}
 \item \textbf{No weighting (no wgt).}
 \item \textbf{TF-IDF,} which weights words with low document frequency higher. We compute TF-IDF scores for each word within each category, therefore calculating the importance of the word in each category.
 \item \textbf{KLD,} which determines the saliency of a word in a category with respect to the rest of the categories. Again KLD leads to a score for each word in each category.
 \item \textbf{TF-CR,} our weighting scheme defined in \S \ref{sec:tfcr}.
\end{compactitem}

\subsection{Varying Sizes of Training Sets}

While we have up to 90,000 instances available as training data, we perform experiments with varying numbers of training instances. This allows us to assess the extent to which weighting schemes can help with varying sizes of training data, provided that calculations of weights using these schemes are done solely from the training data available in each case. We performed experiments for training sets ranging from 1,000 to 9,000. Training instances are randomly sampled in each training scenario, keeping the random sample consistent across different experiments with the same training size, and incrementally adding instances, i.e. a training set of 5,000 contains all of the training instances of that with 4,000 plus another 1,000. All performance scores reported are the result of averaging 10-fold cross-validation experiments.

\section{Results}

\begin{table}[htb]
  \centering
  \small
  \begin{tabular}{l c c c c c c c c}
   \hline
    & \textbf{20ng} & \textbf{hs} & \textbf{ns200} & \textbf{odp} & \textbf{rl} & \textbf{rest} & \textbf{sem} & \textbf{sent} \\
   \hline
   \hline
   \multicolumn{9}{c}{1K training instances} \\
   \hline
   \hline
   no wgt & 0.554 & \textbf{0.613} & 0.448 & \textbf{0.234} & \textbf{0.324} & 0.415 & \textbf{0.577} & \textbf{0.694} \\
   TF-IDF & 0.653 & 0.534 & 0.436 & 0.184 & 0.265 & 0.403 & 0.465 & 0.634 \\
   KLD    & 0.668 & 0.543 & 0.440 & 0.179 & 0.264 & 0.400 & 0.452 & 0.620 \\
   TF-CR  & \textbf{0.783} & 0.566 & \textbf{0.456} & 0.196 & 0.268 & \textbf{0.426} & 0.474 & 0.665 \\
   \hline
   \hline
   \multicolumn{9}{c}{2K training instances} \\
   \hline
   \hline
   no wgt & 0.591 & \textbf{0.620} & 0.473 & \textbf{0.264} & \textbf{0.333} & 0.443 & \textbf{0.592} & \textbf{0.698} \\
   TF-IDF & 0.730 & 0.558 & 0.460 & 0.217 & 0.274 & 0.410 & 0.483 & 0.636 \\
   KLD    & 0.734 & 0.578 & 0.463 & 0.209 & 0.263 & 0.431 & 0.489 & 0.631 \\
   TF-CR  & \textbf{0.836} & 0.588 & \textbf{0.481} & 0.239 & 0.278 & \textbf{0.452} & 0.512 & 0.690 \\
   \hline
   \hline
   \multicolumn{9}{c}{5K training instances} \\
   \hline
   \hline
   no wgt & 0.626 & \textbf{0.634} & 0.503 & 0.290 & \textbf{0.364} & 0.460 & \textbf{0.606} & 0.707 \\
   TF-IDF & 0.645 & 0.545 & 0.476 & 0.266 & 0.308 & 0.408 & 0.503 & 0.633 \\
   KLD    & 0.646 & 0.590 & 0.491 & 0.262 & 0.296 & 0.459 & 0.537 & 0.642 \\
   TF-CR  & \textbf{0.811} & 0.600 & \textbf{0.516} & \textbf{0.296} & 0.332 & \textbf{0.479} & 0.562 & \textbf{0.708} \\
   \hline
   \hline
   \multicolumn{9}{c}{10K training instances} \\
   \hline
   \hline
   no wgt & 0.596 & \textbf{0.641} & 0.519 & 0.301 & \textbf{0.370} & 0.475 & \textbf{0.612} & 0.713 \\
   TF-IDF & 0.440 & 0.536 & 0.490 & 0.295 & 0.303 & 0.399 & 0.502 & 0.636 \\
   KLD    & 0.475 & 0.606 & 0.524 & 0.300 & 0.318 & 0.475 & 0.549 & 0.651 \\
   TF-CR  & \textbf{0.696} & 0.614 & \textbf{0.541} & \textbf{0.351} & 0.347 & \textbf{0.490} & 0.585 & \textbf{0.716} \\
   \hline
   \hline
   \multicolumn{9}{c}{40K training instances} \\
   \hline
   \hline
   no wgt & 0.705 & \textbf{0.656} & 0.538 & 0.312 & 0.418 & 0.495 & \textbf{0.634} & 0.718 \\
   TF-IDF & 0.893 & 0.548 & 0.493 & 0.335 & 0.367 & 0.390 & 0.528 & 0.648 \\
   KLD    & 0.860 & 0.632 & 0.570 & 0.349 & 0.368 & 0.505 & 0.577 & 0.660 \\
   TF-CR  & \textbf{0.930} & 0.635 & \textbf{0.575} & \textbf{0.423} & \textbf{0.427} & \textbf{0.513} & 0.632 & \textbf{0.736} \\
   \hline
   \hline
   \multicolumn{9}{c}{90K training instances} \\
   \hline
   \hline
   no wgt & 0.705 & \textbf{0.661} & 0.544 & 0.325 & 0.422 & 0.503 & 0.635 & 0.721 \\
   TF-IDF & 0.893 & 0.556 & 0.507 & 0.354 & 0.379 & 0.390 & 0.532 & 0.647 \\
   KLD    & 0.860 & 0.643 & 0.586 & 0.362 & 0.364 & 0.516 & 0.577 & 0.663 \\
   TF-CR  & \textbf{0.930} & 0.648 & \textbf{0.595} & \textbf{0.458} & \textbf{0.444} & \textbf{0.522} & \textbf{0.638} & \textbf{0.748} \\
   \hline
  \end{tabular}
  \caption{Comparison of results using different weighting schemes for varying sizes of training data.}
  \label{tab:results}
\end{table}

Table \ref{tab:results} shows the results with varying numbers of training instances. We observe that TF-CR consistently outperforms the other weighting schemes, TF-IDF and KLD, regardless of the training size. The gap between TF-CR and the other weighting schemes generally becomes larger as the training data increases, showing that TF-CR exploits the class distributions more effectively. We can also observe that the unweighted approach outperforms TF-CR in six out of eight datasets when the training data is as small as 1,000 instances. However, TF-CR becomes more effective as the training data increases. TF-CR outperforms the unweighted method in five out of eight datasets with 10,000 training instances, and in seven out of eight datasets with 90,000 training instances. This reinforces the effectiveness of TF-CR for mid-sized training sets and above.

\begin{figure}[htb]
 \centering
 \includegraphics[width=0.4\textwidth]{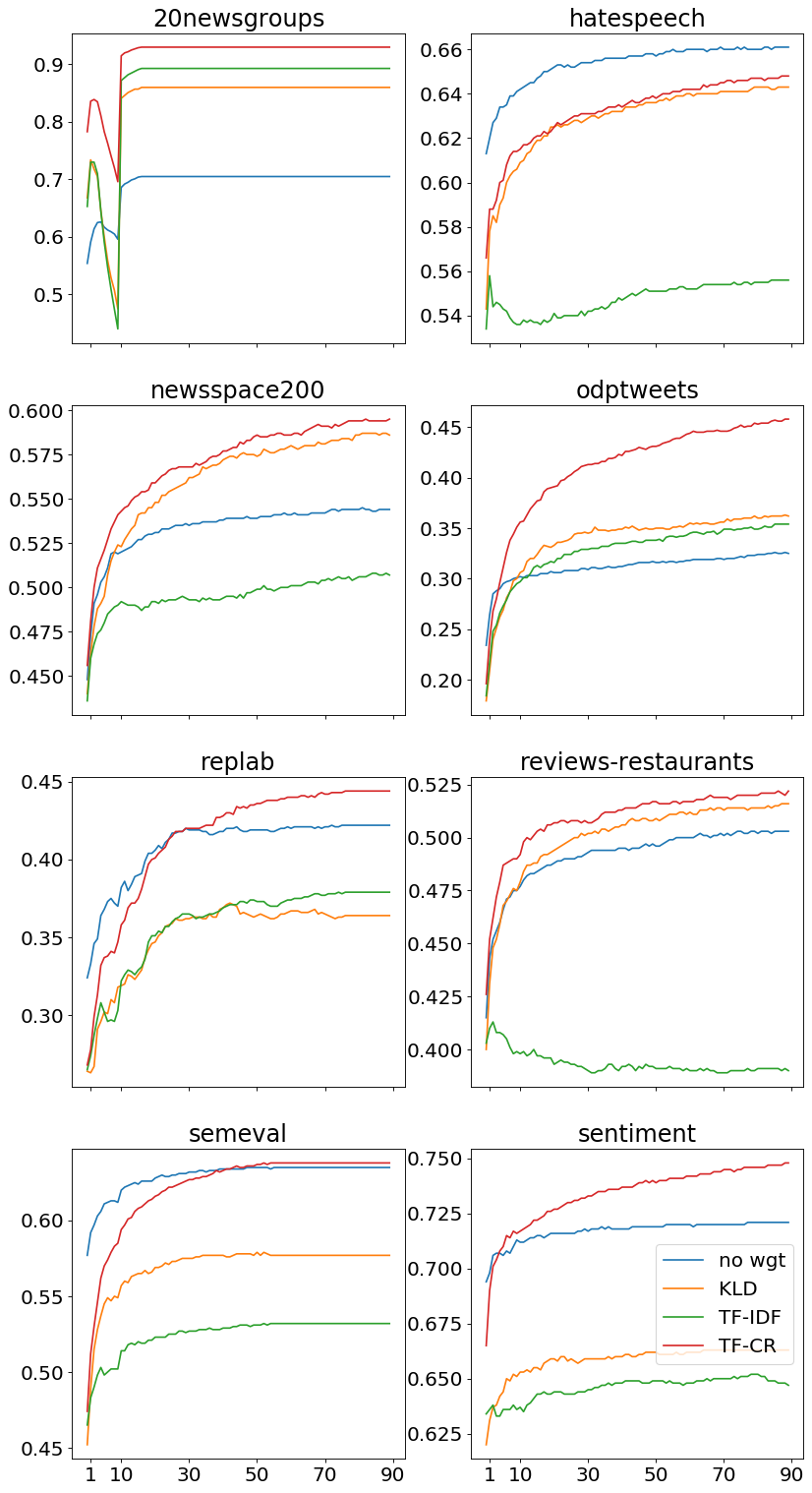}
 \caption{Macro-F1 performance scores for the eight datasets under study, with varying sizes of training data.}
 \label{fig:datasets}
\end{figure}

Figure \ref{fig:datasets} shows the tendency of all four methods as the training size varies from 1,000 to 90,000, with steps of 1,000. With the exception of the \textit{hate speech} dataset, TF-CR outperforms all other methods for larger training sets. Moreover, TF-CR consistently outperforms all other methods for most training sizes in five datasets: \textit{20newsgroups}, \textit{newsspace200}, \textit{odptweets}, \textit{restaurants} and \textit{sentiment}.

\section{Discussion}

We have introduced TF-CR,\footnote{Code available at \url{https://github.com/azubiaga/tfcr}} a first-of-its-kind weighting scheme that can leverage word distributions across categories in training data for text classification. The intuition behind TF-CR is to give higher weights to frequent words that exclusively or predominantly occur within a category. This leads to category-specific weights for each word, which allows an embedding representation that captures varying importances of words across categories. Experimenting on eight datasets, we show that (1) it improves over unweighted word embeddings in seven of the datasets with large training datasets, (2) it improves consistently for most training sizes in five of the datasets. TF-CR also consistently outperforms TF-IDF and KLD.

Our objective here has been to introduce and validate TF-CR. Additional tuning of classifier parameters, adding features, etc. for achieving state-of-the-art performance is beyond the scope of this work. We also aim to extend this work by further exploring the differences across datasets, to determine dataset characteristics that maximise the benefits of TF-CR.

\section*{Acknowledgments}

This research utilised Queen Mary's Apocrita HPC facility, supported by QMUL Research-IT. http://doi.org/10.5281/zenodo.438045

\bibliographystyle{ACM-Reference-Format}
\bibliography{tfcr}


\begin{thebibliography}{31}


\ifx \showCODEN    \undefined \def \showCODEN     #1{\unskip}     \fi
\ifx \showDOI      \undefined \def \showDOI       #1{#1}\fi
\ifx \showISBNx    \undefined \def \showISBNx     #1{\unskip}     \fi
\ifx \showISBNxiii \undefined \def \showISBNxiii  #1{\unskip}     \fi
\ifx \showISSN     \undefined \def \showISSN      #1{\unskip}     \fi
\ifx \showLCCN     \undefined \def \showLCCN      #1{\unskip}     \fi
\ifx \shownote     \undefined \def \shownote      #1{#1}          \fi
\ifx \showarticletitle \undefined \def \showarticletitle #1{#1}   \fi
\ifx \showURL      \undefined \def \showURL       {\relax}        \fi
\providecommand\bibfield[2]{#2}
\providecommand\bibinfo[2]{#2}
\providecommand\natexlab[1]{#1}
\providecommand\showeprint[2][]{arXiv:#2}

\bibitem[\protect\citeauthoryear{Amig{\'o}, {Carrillo de Albornoz}, Chugur,
  Corujo, Gonzalo, Mart{\'i}n, Meij, de~Rijke, and Spina}{Amig{\'o}
  et~al\mbox{.}}{2013}]%
        {replab2013overview}
\bibfield{author}{\bibinfo{person}{E. Amig{\'o}}, \bibinfo{person}{J. {Carrillo
  de Albornoz}}, \bibinfo{person}{I. Chugur}, \bibinfo{person}{A. Corujo},
  \bibinfo{person}{J. Gonzalo}, \bibinfo{person}{T. Mart{\'i}n},
  \bibinfo{person}{E. Meij}, \bibinfo{person}{M. de Rijke}, {and}
  \bibinfo{person}{D. Spina}.} \bibinfo{year}{2013}\natexlab{}.
\newblock \showarticletitle{{Overview of RepLab 2013: Evaluating Online
  Reputation Monitoring Systems}}. In \bibinfo{booktitle}{\emph{{Proceedings of
  CLEF}}}. \bibinfo{pages}{333--352}.
\newblock


\bibitem[\protect\citeauthoryear{Bengio, Ducharme, Vincent, and Jauvin}{Bengio
  et~al\mbox{.}}{2003}]%
        {bengio2003neural}
\bibfield{author}{\bibinfo{person}{Yoshua Bengio}, \bibinfo{person}{R{\'e}jean
  Ducharme}, \bibinfo{person}{Pascal Vincent}, {and} \bibinfo{person}{Christian
  Jauvin}.} \bibinfo{year}{2003}\natexlab{}.
\newblock \showarticletitle{A neural probabilistic language model}.
\newblock \bibinfo{journal}{\emph{Journal of machine learning research}}
  \bibinfo{volume}{3}, \bibinfo{number}{Feb} (\bibinfo{year}{2003}),
  \bibinfo{pages}{1137--1155}.
\newblock


\bibitem[\protect\citeauthoryear{Bollegala, Mu, and Goulermas}{Bollegala
  et~al\mbox{.}}{2016}]%
        {bollegala2016cross}
\bibfield{author}{\bibinfo{person}{Danushka Bollegala},
  \bibinfo{person}{Tingting Mu}, {and} \bibinfo{person}{John~Yannis
  Goulermas}.} \bibinfo{year}{2016}\natexlab{}.
\newblock \showarticletitle{Cross-domain sentiment classification using
  sentiment sensitive embeddings}.
\newblock \bibinfo{journal}{\emph{IEEE Transactions on Knowledge and Data
  Engineering}} \bibinfo{volume}{28}, \bibinfo{number}{2}
  (\bibinfo{year}{2016}), \bibinfo{pages}{398--410}.
\newblock


\bibitem[\protect\citeauthoryear{Chiu, Crichton, Korhonen, and Pyysalo}{Chiu
  et~al\mbox{.}}{2016}]%
        {chiu2016train}
\bibfield{author}{\bibinfo{person}{Billy Chiu}, \bibinfo{person}{Gamal
  Crichton}, \bibinfo{person}{Anna Korhonen}, {and} \bibinfo{person}{Sampo
  Pyysalo}.} \bibinfo{year}{2016}\natexlab{}.
\newblock \showarticletitle{How to train good word embeddings for biomedical
  NLP}. In \bibinfo{booktitle}{\emph{Proceedings of the 15th Workshop on
  Biomedical Natural Language Processing}}. \bibinfo{pages}{166--174}.
\newblock


\bibitem[\protect\citeauthoryear{Cortis, Freitas, Daudert, Huerlimann, Zarrouk,
  Handschuh, and Davis}{Cortis et~al\mbox{.}}{2017}]%
        {cortis2017semeval}
\bibfield{author}{\bibinfo{person}{Keith Cortis}, \bibinfo{person}{Andr{\'e}
  Freitas}, \bibinfo{person}{Tobias Daudert}, \bibinfo{person}{Manuela
  Huerlimann}, \bibinfo{person}{Manel Zarrouk}, \bibinfo{person}{Siegfried
  Handschuh}, {and} \bibinfo{person}{Brian Davis}.}
  \bibinfo{year}{2017}\natexlab{}.
\newblock \showarticletitle{Semeval-2017 task 5: Fine-grained sentiment
  analysis on financial microblogs and news}. In
  \bibinfo{booktitle}{\emph{Proceedings of SemEval}}.
  \bibinfo{pages}{519--535}.
\newblock


\bibitem[\protect\citeauthoryear{Del~Corso, Gulli, and Romani}{Del~Corso
  et~al\mbox{.}}{2005}]%
        {del2005ranking}
\bibfield{author}{\bibinfo{person}{Gianna~M Del~Corso},
  \bibinfo{person}{Antonio Gulli}, {and} \bibinfo{person}{Francesco Romani}.}
  \bibinfo{year}{2005}\natexlab{}.
\newblock \showarticletitle{Ranking a stream of news}. In
  \bibinfo{booktitle}{\emph{Proceedings of WWW}}. ACM,
  \bibinfo{pages}{97--106}.
\newblock


\bibitem[\protect\citeauthoryear{Founta, Djouvas, Chatzakou, Leontiadis,
  Blackburn, Stringhini, Vakali, Sirivianos, and Kourtellis}{Founta
  et~al\mbox{.}}{2018}]%
        {founta2018large}
\bibfield{author}{\bibinfo{person}{Antigoni-Maria Founta},
  \bibinfo{person}{Constantinos Djouvas}, \bibinfo{person}{Despoina Chatzakou},
  \bibinfo{person}{Ilias Leontiadis}, \bibinfo{person}{Jeremy Blackburn},
  \bibinfo{person}{Gianluca Stringhini}, \bibinfo{person}{Athena Vakali},
  \bibinfo{person}{Michael Sirivianos}, {and} \bibinfo{person}{Nicolas
  Kourtellis}.} \bibinfo{year}{2018}\natexlab{}.
\newblock \showarticletitle{Large Scale Crowdsourcing and Characterization of
  Twitter Abusive Behavior}. In \bibinfo{booktitle}{\emph{Proceedings of
  ICWSM}}. AAAI Press.
\newblock


\bibitem[\protect\citeauthoryear{Ganguly, Roy, Mitra, and Jones}{Ganguly
  et~al\mbox{.}}{2015}]%
        {ganguly2015word}
\bibfield{author}{\bibinfo{person}{Debasis Ganguly}, \bibinfo{person}{Dwaipayan
  Roy}, \bibinfo{person}{Mandar Mitra}, {and} \bibinfo{person}{Gareth~JF
  Jones}.} \bibinfo{year}{2015}\natexlab{}.
\newblock \showarticletitle{Word embedding based generalized language model for
  information retrieval}. In \bibinfo{booktitle}{\emph{Proceedings of SIGIR}}.
  \bibinfo{pages}{795--798}.
\newblock


\bibitem[\protect\citeauthoryear{Go, Bhayani, and Huang}{Go
  et~al\mbox{.}}{2009}]%
        {go2009twitter}
\bibfield{author}{\bibinfo{person}{Alec Go}, \bibinfo{person}{Richa Bhayani},
  {and} \bibinfo{person}{Lei Huang}.} \bibinfo{year}{2009}\natexlab{}.
\newblock \showarticletitle{Twitter sentiment classification using distant
  supervision}.
\newblock \bibinfo{journal}{\emph{CS224N Project Report, Stanford}}
  \bibinfo{volume}{1}, \bibinfo{number}{12} (\bibinfo{year}{2009}).
\newblock


\bibitem[\protect\citeauthoryear{Godin, Vandersmissen, De~Neve, and Van~de
  Walle}{Godin et~al\mbox{.}}{2015}]%
        {godin2015multimedia}
\bibfield{author}{\bibinfo{person}{Fr{\'e}deric Godin},
  \bibinfo{person}{Baptist Vandersmissen}, \bibinfo{person}{Wesley De~Neve},
  {and} \bibinfo{person}{Rik Van~de Walle}.} \bibinfo{year}{2015}\natexlab{}.
\newblock \showarticletitle{Multimedia Lab $@ $ ACL WNUT NER Shared Task: Named
  Entity Recognition for Twitter Microposts using Distributed Word
  Representations}. In \bibinfo{booktitle}{\emph{Proceedings of the Workshop on
  Noisy User-generated Text}}. \bibinfo{pages}{146--153}.
\newblock


\bibitem[\protect\citeauthoryear{Grohe}{Grohe}{2020}]%
        {grohe2020word2vec}
\bibfield{author}{\bibinfo{person}{Martin Grohe}.}
  \bibinfo{year}{2020}\natexlab{}.
\newblock \showarticletitle{word2vec, node2vec, graph2vec, X2vec: Towards a
  Theory of Vector Embeddings of Structured Data}.
\newblock \bibinfo{journal}{\emph{arXiv preprint arXiv:2003.12590}}
  (\bibinfo{year}{2020}).
\newblock


\bibitem[\protect\citeauthoryear{Jiang and Zubiaga}{Jiang and Zubiaga}{2019}]%
        {jiang2019leveraging}
\bibfield{author}{\bibinfo{person}{Aiqi Jiang} {and} \bibinfo{person}{Arkaitz
  Zubiaga}.} \bibinfo{year}{2019}\natexlab{}.
\newblock \showarticletitle{Leveraging aspect phrase embeddings for
  cross-domain review rating prediction}.
\newblock \bibinfo{journal}{\emph{PeerJ Computer Science}}  \bibinfo{volume}{5}
  (\bibinfo{year}{2019}), \bibinfo{pages}{e225}.
\newblock


\bibitem[\protect\citeauthoryear{Jones}{Jones}{1972}]%
        {jones1972statistical}
\bibfield{author}{\bibinfo{person}{Karen~Sp{\"a}rck Jones}.}
  \bibinfo{year}{1972}\natexlab{}.
\newblock \showarticletitle{A statistical interpretation of term specificity
  and its application in retrieval}.
\newblock \bibinfo{journal}{\emph{Journal of documentation}}
  \bibinfo{volume}{28} (\bibinfo{year}{1972}), \bibinfo{pages}{11--21}.
\newblock


\bibitem[\protect\citeauthoryear{Kuang and Davison}{Kuang and Davison}{2018}]%
        {kuang2018class}
\bibfield{author}{\bibinfo{person}{Sicong Kuang} {and} \bibinfo{person}{Brian~D
  Davison}.} \bibinfo{year}{2018}\natexlab{}.
\newblock \showarticletitle{Class-specific word embedding through linear
  compositionality}. In \bibinfo{booktitle}{\emph{2018 IEEE International
  Conference on Big Data and Smart Computing (BigComp)}}. IEEE,
  \bibinfo{pages}{390--397}.
\newblock


\bibitem[\protect\citeauthoryear{Kuang and Davison}{Kuang and Davison}{2019}]%
        {kuang2019learning}
\bibfield{author}{\bibinfo{person}{Sicong Kuang} {and} \bibinfo{person}{Brian~D
  Davison}.} \bibinfo{year}{2019}\natexlab{}.
\newblock \showarticletitle{Learning class-specific word embeddings}.
\newblock \bibinfo{journal}{\emph{The Journal of Supercomputing}}
  (\bibinfo{year}{2019}), \bibinfo{pages}{1--28}.
\newblock


\bibitem[\protect\citeauthoryear{Kullback and Leibler}{Kullback and
  Leibler}{1951}]%
        {kullback1951information}
\bibfield{author}{\bibinfo{person}{Solomon Kullback} {and}
  \bibinfo{person}{Richard~A Leibler}.} \bibinfo{year}{1951}\natexlab{}.
\newblock \showarticletitle{On information and sufficiency}.
\newblock \bibinfo{journal}{\emph{The annals of mathematical statistics}}
  \bibinfo{volume}{22}, \bibinfo{number}{1} (\bibinfo{year}{1951}),
  \bibinfo{pages}{79--86}.
\newblock


\bibitem[\protect\citeauthoryear{Mikolov, Sutskever, Chen, Corrado, and
  Dean}{Mikolov et~al\mbox{.}}{2013}]%
        {mikolov2013distributed}
\bibfield{author}{\bibinfo{person}{Tomas Mikolov}, \bibinfo{person}{Ilya
  Sutskever}, \bibinfo{person}{Kai Chen}, \bibinfo{person}{Greg~S Corrado},
  {and} \bibinfo{person}{Jeff Dean}.} \bibinfo{year}{2013}\natexlab{}.
\newblock \showarticletitle{Distributed representations of words and phrases
  and their compositionality}. In \bibinfo{booktitle}{\emph{Advances in neural
  information processing systems}}. \bibinfo{pages}{3111--3119}.
\newblock


\bibitem[\protect\citeauthoryear{Musto, Semeraro, de~Gemmis, and Lops}{Musto
  et~al\mbox{.}}{2016}]%
        {musto2016learning}
\bibfield{author}{\bibinfo{person}{Cataldo Musto}, \bibinfo{person}{Giovanni
  Semeraro}, \bibinfo{person}{Marco de Gemmis}, {and} \bibinfo{person}{Pasquale
  Lops}.} \bibinfo{year}{2016}\natexlab{}.
\newblock \showarticletitle{Learning word embeddings from wikipedia for
  content-based recommender systems}. In \bibinfo{booktitle}{\emph{European
  Conference on Information Retrieval}}. Springer, \bibinfo{pages}{729--734}.
\newblock


\bibitem[\protect\citeauthoryear{Pennington, Socher, and Manning}{Pennington
  et~al\mbox{.}}{2014}]%
        {pennington2014glove}
\bibfield{author}{\bibinfo{person}{Jeffrey Pennington},
  \bibinfo{person}{Richard Socher}, {and} \bibinfo{person}{Christopher
  Manning}.} \bibinfo{year}{2014}\natexlab{}.
\newblock \showarticletitle{Glove: Global vectors for word representation}. In
  \bibinfo{booktitle}{\emph{Proceedings of EMNLP}}.
  \bibinfo{pages}{1532--1543}.
\newblock


\bibitem[\protect\citeauthoryear{Pilehvar and Camacho-Collados}{Pilehvar and
  Camacho-Collados}{2020}]%
        {pilehvar2020embeddings}
\bibfield{author}{\bibinfo{person}{Mohammad~Taher Pilehvar} {and}
  \bibinfo{person}{Jose Camacho-Collados}.} \bibinfo{year}{2020}\natexlab{}.
\newblock \bibinfo{booktitle}{\emph{Embeddings in Natural Language Processing:
  Theory and Advances in Vector Representation of Meaning}}.
\newblock \bibinfo{publisher}{Morgan \& Claypool}.
\newblock


\bibitem[\protect\citeauthoryear{Rosenthal, Farra, and Nakov}{Rosenthal
  et~al\mbox{.}}{2017}]%
        {rosenthal2017semeval}
\bibfield{author}{\bibinfo{person}{Sara Rosenthal}, \bibinfo{person}{Noura
  Farra}, {and} \bibinfo{person}{Preslav Nakov}.}
  \bibinfo{year}{2017}\natexlab{}.
\newblock \showarticletitle{SemEval-2017 task 4: Sentiment analysis in
  Twitter}. In \bibinfo{booktitle}{\emph{Proceedings of SemEval}}.
  \bibinfo{pages}{502--518}.
\newblock


\bibitem[\protect\citeauthoryear{Salton and Buckley}{Salton and
  Buckley}{1988}]%
        {salton1988term}
\bibfield{author}{\bibinfo{person}{Gerard Salton} {and}
  \bibinfo{person}{Christopher Buckley}.} \bibinfo{year}{1988}\natexlab{}.
\newblock \showarticletitle{Term-weighting approaches in automatic text
  retrieval}.
\newblock \bibinfo{journal}{\emph{Information processing \& management}}
  \bibinfo{volume}{24}, \bibinfo{number}{5} (\bibinfo{year}{1988}),
  \bibinfo{pages}{513--523}.
\newblock


\bibitem[\protect\citeauthoryear{Sebastiani}{Sebastiani}{2002}]%
        {sebastiani2002machine}
\bibfield{author}{\bibinfo{person}{Fabrizio Sebastiani}.}
  \bibinfo{year}{2002}\natexlab{}.
\newblock \showarticletitle{Machine learning in automated text categorization}.
\newblock \bibinfo{journal}{\emph{ACM computing surveys (CSUR)}}
  \bibinfo{volume}{34}, \bibinfo{number}{1} (\bibinfo{year}{2002}),
  \bibinfo{pages}{1--47}.
\newblock


\bibitem[\protect\citeauthoryear{Tang}{Tang}{2015}]%
        {tang2015sentiment}
\bibfield{author}{\bibinfo{person}{Duyu Tang}.}
  \bibinfo{year}{2015}\natexlab{}.
\newblock \showarticletitle{Sentiment-specific representation learning for
  document-level sentiment analysis}. In \bibinfo{booktitle}{\emph{Proceedings
  of WSDM}}. ACM, \bibinfo{pages}{447--452}.
\newblock


\bibitem[\protect\citeauthoryear{Tang, Wei, Qin, Yang, Liu, and Zhou}{Tang
  et~al\mbox{.}}{2016}]%
        {tang2016sentiment}
\bibfield{author}{\bibinfo{person}{Duyu Tang}, \bibinfo{person}{Furu Wei},
  \bibinfo{person}{Bing Qin}, \bibinfo{person}{Nan Yang}, \bibinfo{person}{Ting
  Liu}, {and} \bibinfo{person}{Ming Zhou}.} \bibinfo{year}{2016}\natexlab{}.
\newblock \showarticletitle{Sentiment embeddings with applications to sentiment
  analysis}.
\newblock \bibinfo{journal}{\emph{IEEE Transactions on Knowledge and Data
  Engineering}} \bibinfo{volume}{28}, \bibinfo{number}{2}
  (\bibinfo{year}{2016}), \bibinfo{pages}{496--509}.
\newblock


\bibitem[\protect\citeauthoryear{Tang, Wei, Yang, Zhou, Liu, and Qin}{Tang
  et~al\mbox{.}}{2014}]%
        {tang2014learning}
\bibfield{author}{\bibinfo{person}{Duyu Tang}, \bibinfo{person}{Furu Wei},
  \bibinfo{person}{Nan Yang}, \bibinfo{person}{Ming Zhou},
  \bibinfo{person}{Ting Liu}, {and} \bibinfo{person}{Bing Qin}.}
  \bibinfo{year}{2014}\natexlab{}.
\newblock \showarticletitle{Learning sentiment-specific word embedding for
  twitter sentiment classification}. In \bibinfo{booktitle}{\emph{Proceedings
  of ACL}}, Vol.~\bibinfo{volume}{1}. \bibinfo{pages}{1555--1565}.
\newblock


\bibitem[\protect\citeauthoryear{Wang, Xu, Xu, Tian, Liu, and Hao}{Wang
  et~al\mbox{.}}{2016}]%
        {wang2016semantic}
\bibfield{author}{\bibinfo{person}{Peng Wang}, \bibinfo{person}{Bo Xu},
  \bibinfo{person}{Jiaming Xu}, \bibinfo{person}{Guanhua Tian},
  \bibinfo{person}{Cheng-Lin Liu}, {and} \bibinfo{person}{Hongwei Hao}.}
  \bibinfo{year}{2016}\natexlab{}.
\newblock \showarticletitle{Semantic expansion using word embedding clustering
  and convolutional neural network for improving short text classification}.
\newblock \bibinfo{journal}{\emph{Neurocomputing}}  \bibinfo{volume}{174}
  (\bibinfo{year}{2016}), \bibinfo{pages}{806--814}.
\newblock


\bibitem[\protect\citeauthoryear{Zhang, Guo, Lan, Xu, and Cheng}{Zhang
  et~al\mbox{.}}{2018}]%
        {zhang2018aggregating}
\bibfield{author}{\bibinfo{person}{Ruqing Zhang}, \bibinfo{person}{Jiafeng
  Guo}, \bibinfo{person}{Yanyan Lan}, \bibinfo{person}{Jun Xu}, {and}
  \bibinfo{person}{Xueqi Cheng}.} \bibinfo{year}{2018}\natexlab{}.
\newblock \showarticletitle{Aggregating Neural Word Embeddings for Document
  Representation}. In \bibinfo{booktitle}{\emph{Proceedings of ECIR}}.
  Springer, \bibinfo{pages}{303--315}.
\newblock


\bibitem[\protect\citeauthoryear{Zou, Socher, Cer, and Manning}{Zou
  et~al\mbox{.}}{2013}]%
        {zou2013bilingual}
\bibfield{author}{\bibinfo{person}{Will~Y Zou}, \bibinfo{person}{Richard
  Socher}, \bibinfo{person}{Daniel Cer}, {and} \bibinfo{person}{Christopher~D
  Manning}.} \bibinfo{year}{2013}\natexlab{}.
\newblock \showarticletitle{Bilingual word embeddings for phrase-based machine
  translation}. In \bibinfo{booktitle}{\emph{Proceedings of EMNLP}}.
  \bibinfo{pages}{1393--1398}.
\newblock


\bibitem[\protect\citeauthoryear{Zubiaga and Ji}{Zubiaga and Ji}{2013}]%
        {zubiaga2013harnessing}
\bibfield{author}{\bibinfo{person}{Arkaitz Zubiaga} {and} \bibinfo{person}{Heng
  Ji}.} \bibinfo{year}{2013}\natexlab{}.
\newblock \showarticletitle{Harnessing web page directories for large-scale
  classification of tweets}. In \bibinfo{booktitle}{\emph{Proceedings of WWW}}.
  ACM, \bibinfo{pages}{225--226}.
\newblock


\bibitem[\protect\citeauthoryear{Zubiaga and Jiang}{Zubiaga and Jiang}{2020}]%
        {zubiaga2018learning}
\bibfield{author}{\bibinfo{person}{Arkaitz Zubiaga} {and} \bibinfo{person}{Aiqi
  Jiang}.} \bibinfo{year}{2020}\natexlab{}.
\newblock \showarticletitle{Early Detection of Social Media Hoaxes at Scale}.
\newblock \bibinfo{journal}{\emph{ACM Transactions on the Web (TWEB)}}
  \bibinfo{volume}{14}, \bibinfo{number}{4} (\bibinfo{year}{2020}),
  \bibinfo{pages}{1--23}.
\newblock


\end{thebibliography}

\end{document}